\documentclass[letterpaper, 10 pt, conference]{ieeeconf}  

\IEEEoverridecommandlockouts                              

\usepackage{times} 
\usepackage{amsmath} 
\usepackage{amssymb}  
\usepackage{subcaption}
\usepackage{graphicx}
\usepackage{algorithm}
\usepackage{dsfont}
\usepackage{multirow}
\usepackage[noend]{algpseudocode}
\usepackage{adjustbox}
\usepackage{booktabs}
\usepackage{amssymb}
\usepackage{pifont}
\newcommand{\cmark}{\ding{51}}%
\newcommand{\xmark}{\ding{55}}%

\usepackage{cancel}
\usepackage{xcolor}

\def\A{\mathcal{A}}
\def\P{\mathcal{P}}
\def\S{\mathcal{S}}
\def\J{\mathcal{J}}

\def\L{\mathcal{L}}
\def\piPHI{\pi_\phi(s)}
\def\piPHII{\pi_{\phi'}(s)}
\def\logpi{\log \pi_\phi(a|s)}

\def\Qr{Q_{\omega_r}}
\def\Qc{Q_{\omega_c}}
\newcommand\scalemath[2]{\scalebox{#1}{\mbox{\ensuremath{\displaystyle #2}}}}

\newcommand{\colorcancelto}[2]{\renewcommand\CancelColor{\color{#1}}\cancelto{#2}}

\let\labelindent\relax
\usepackage{enumitem}
\setitemize{leftmargin=3.5mm}

\title{\LARGE \bf
Meta SAC-Lag: Towards Deployable Safe Reinforcement Learning via MetaGradient-based Hyperparameter Tuning
}

\author{Homayoun Honari$^{1}$, Amir M. Soufi Enayati$^{1}$, Mehran Ghafarian Tamizi$^{2}$, Homayoun Najjaran$^{1,2}$
\thanks{*This work was supported by Apera AI and Mathematics of Information Technology and Complex Systems (MITACS) under IT16412 Mitacs Accelerate and Natural Sciences and Engineering Research Council (NSERC) Canada under the Grant RTI-2023-00418, and a partial equipment support was received from Kinova\textregistered~Inc.}
\thanks{$^{1}$Department of Mechanical Engineering, University of Victoria, Victoria, BC, Canada
        {\tt\small \{hmnhonari,amsoufi,najjaran\}@uvic.ca}}%
\thanks{$^{2}$Department of Electrical and Computer Engineering, University of Victoria, Victoria, BC, Canada
        {\tt\small mehranght@uvic.ca}}%
}

\begin{document}

\maketitle
\thispagestyle{empty}
\pagestyle{empty}

\begin{abstract}

Safe Reinforcement Learning (Safe RL) is one of the prevalently studied subcategories of trial-and-error-based methods with the intention to be deployed on real-world systems. 
In safe RL, the goal is to maximize reward performance while minimizing constraints, often achieved by setting bounds on constraint functions and utilizing the Lagrangian method. However, deploying Lagrangian-based safe RL in real-world scenarios is challenging due to the necessity of threshold fine-tuning, as imprecise adjustments may lead to suboptimal policy convergence.
To mitigate this challenge, we propose a unified Lagrangian-based model-free architecture called \textit{Meta Soft Actor-Critic Lagrangian} (Meta SAC-Lag). Meta SAC-Lag uses meta-gradient optimization to automatically update the safety-related hyperparameters. The proposed method is designed to address safe exploration and threshold adjustment with minimal hyperparameter tuning requirement. In our pipeline, the inner parameters are updated through the conventional formulation and the hyperparameters are adjusted using the meta-objectives which are defined based on the updated parameters. Our results show that the agent can reliably adjust the safety performance due to the relatively fast convergence rate of the safety threshold.
We evaluate the performance of Meta SAC-Lag in five simulated environments against Lagrangian baselines, and the results demonstrate its capability to create synergy between parameters, yielding better or competitive results. Furthermore, we conduct a real-world experiment involving a robotic arm tasked with pouring coffee into a cup without spillage. Meta SAC-Lag is successfully trained to execute the task, while minimizing effort constraints. 
The success of Meta SAC-Lag in performing the experiment is intended to be a step toward practical deployment of safe RL algorithms to learn the control process of safety-critical real-world systems without explicit engineering.


\end{abstract}

\section{Introduction}

Reinforcement Learning (RL) is one of the most important paradigms for learning to control physical systems. However, a major shortcoming of RL is its need for exploration and extensive trial and error. For that reason, while we observe its wide success in various domains such as Energy systems~\cite{PERERA2021110618}, Video games~\cite{shao2019survey}, and Robotics~\cite{andrychowicz2020learning}, the real-world deployment of these algorithms to learn the control process poses is challenging~\cite{dulac2019challenges} since the exploration process might lead the system to states that might damage the system and incur heavy costs to the user. To this end, safe RL methods aim to address this issue by optimizing the policy such that it is compliant with the constraints. The constraints are defined such that they aim to prevent the system from exceeding its physical limitations.

The most common approach in safe RL is through the Lagrangian method. Specified under the Constrained Markov Decision Process (CMDP) framework, through defining thresholds for the constraints, the multi-objective optimization problem is converted to constraint satisfaction and is solved by casting it to an unconstrained problem using the Lagrangian method.
While the approach has been extensively studied in the literature~\cite{gu2022review,brunke2022safe}, without precise tuning and engineering of the constraint thresholds, the Lagrangian methods will suffer from convergence to suboptimal policies. For that reason, the real-world use of these algorithms is rendered to be challenging due to the iterative process of hyperparameter tuning.

\begin{figure}[t]
    \centering
    \captionsetup{justification=centering}
    \includegraphics[trim={3cm 3cm 2cm 0},clip,width=1\columnwidth,height=1\linewidth]{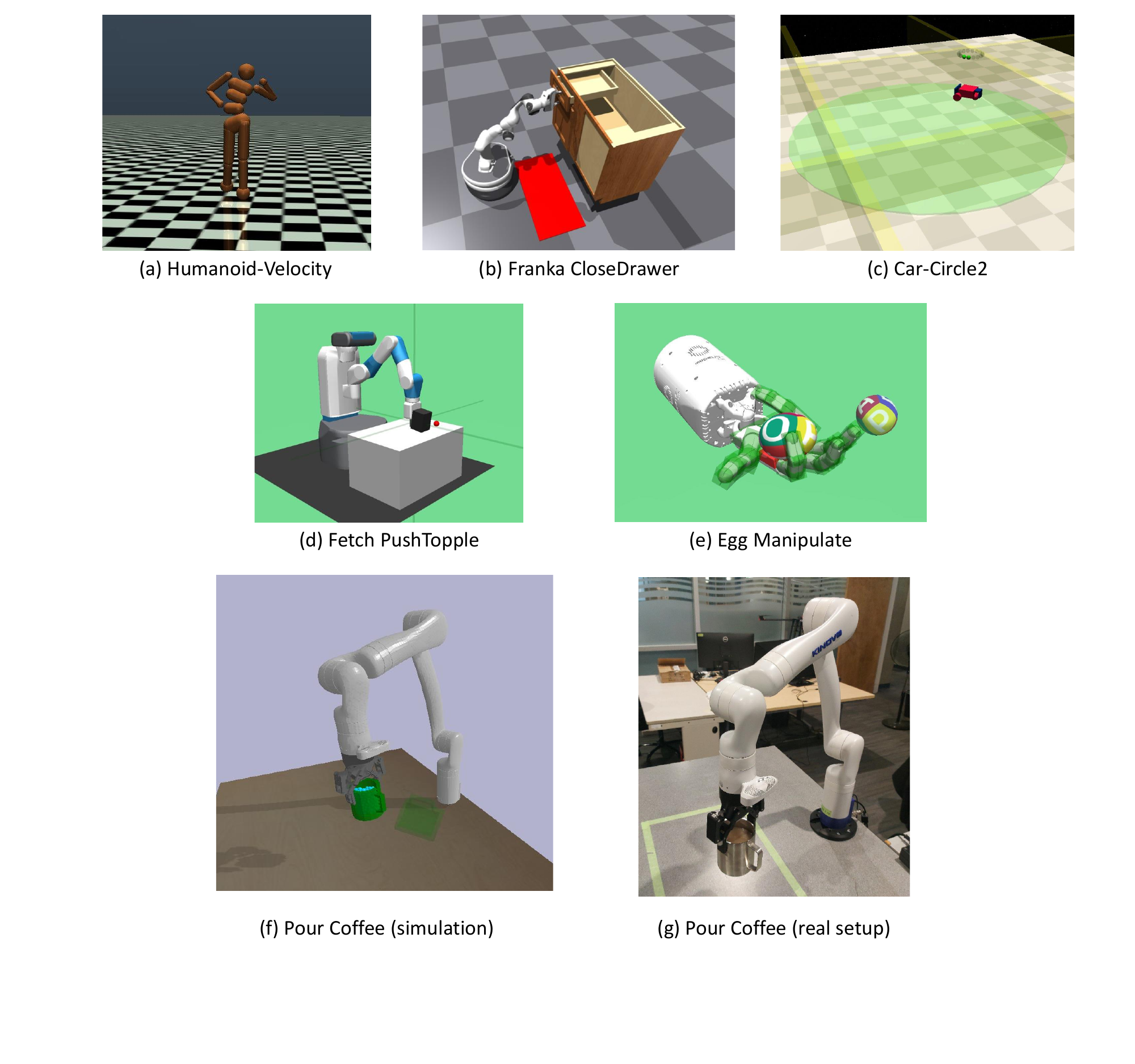}
    \caption{Safety-critical environments used to deploy Meta SAC-Lag. The top two rows represent simulated environments with four general safety topics: locomotion (a), obstacle avoidance (b,c), robotic manipulation (d), dexterous manipulation (e). The bottom row represents Pour Coffee environment (f,g) used to study the deployability of the algorithm in a real-world setup.}
    \label{render}
\vspace{-1.7em}
\end{figure}

To address these challenges, based on the Soft Actor-Critic (SAC) architecture~\cite{haarnoja2018soft}, our algorithm aims to address two fundamental problems: safe exploration and tuning-free constraint adjustment. Previous attempts to automate the tuning of exploration-related hyperparameter of SAC have been mostly focused on optimizing the performance of the system~\cite{haarnoja2019soft,wang2020meta}. However, addressing the safety compliance of SAC has been limited. To this end, we propose a threshold-free safety-aware exploration optimization pipeline. In addition, our approach optimizes the safety threshold according to the overall performance of the policy. We are able to update the aforementioned hyperparameters using the metagradients w.r.t. to the meta-objectives which are computed based on the gradients of the internal learnable parameters. Finally, to assess the performance of Meta SAC-Lag, as depicted in Fig.~\ref{render}, we study its performance against several baseline algorithms in five simulated robotic tasks with four different application themes. We observe that our method attains better or comparable results in terms of safety or reward performance while automatically tuning the safety-related hyperparameters. Furthermore, we present a safety benchmark test case, called \textit{Pour Coffee}, which attempts to relocate and pour a coffee-filled mug into another cup. Constraint violation happens in case of collision or the spillage of the coffee. We deploy and train Meta SAC-Lag in a real-world setup using Kinova Gen3 robot. Our implementation shows that not only is Meta SAC-Lag capable of safe deployment without the iterative process of hyperparameter tuning but also, the learning process of the policy results in a smooth and jerk-free execution of the task with minimum effort imposed on the system.

\noindent Our main contributions can be summarized as:
\begin{itemize}
    \item We propose a Lagrangian-based safe RL method able to automatically adjust the constraint bounds.
    \item Meta SAC-Lag addresses safe exploration through an unconstrained metagradient-based optimization pipeline.
    \item We validate the applicability of Meta SAC-Lag in five simulated robotic environments against baseline algorithms.
    \item A test environment, called \textit{Pour Coffee}, is presented, and, with minimal prior safety-related hyperparameter tuning, Meta SAC-Lag is trained on a real-world Kinova Gen3 setup. The algorithm successfully achieves the task objective with minimized effort exerted on the robot.
\end{itemize}

\section{Related Work}
The constrained Markov decision process (CMDP), is the theoretical building block of safe RL. CMDPs have been widely studied in the RL paradigm~\cite{sutton2018reinforcement, altman2021constrained} and are solved using Lagrangian methods~\cite{bertsekas2012dynamic}. In this regard, Shen et. al~\cite{shen2014} devised the risk-sensitive policy
optimization (RSPO) algorithm which sequentially decreases the Lagrangian multiplier to zero. Furthermore, Stooke et. al~\cite{stooke2020responsive} updates the multiplier using PID control. Additionally, Reward-constrained policy optimization (RCPO) employs dual gradient descent optimization for the policy and Lagrange multiplier~\cite{tessler2018reward}.

In another aspect, metagradient optimization has been explored thoroughly in RL hyperparameter tuning. Initially, model-agnostic meta-learning (MAML)~\cite{finn2017model} introduced meta-optimization of initial weights to enable fast task adaptation within a few gradient descent steps. In a different approach, Meta-Gradient RL~\cite{xu2018meta} extended the concept to learn the hyperparameters of return functions online. This paradigm offered a general approach, applicable to other RL hyperparameters. Subsequently, similar techniques were applied for auto-tuning other RL hyperparameters, such as exploration thresholds~\cite{haarnoja2018learning}, entropy temperature in SAC~\cite{wang2020meta}, auxiliary tasks and sub-goals~\cite{veeriah2021discovery}, and differentiable hyperparameters of loss functions~\cite{zahavy2020self}. 
Despite these advancements, metagradient methods have not been extensively explored in constrained RL paradigms~\cite{gu2022review}, with few applications focusing on ensuring safety in sensitive learning tasks, as seen in the work by Calian et al.~\cite{calian2020balancing}. The authors utilized meta-gradients to update the Lagrange multiplier learning rate in an off-policy RL framework.

The Lagrangian methods are not the sole approach taken toward solving safe RL. Thananjeyan et al.~\cite{thananjeyan2021recovery} trained a recovery policy in parallel to the task policy and used it whenever the task policy chose actions deemed too risky. More prominently, model-based RL and safety guarantees for risk aversion during training are proposed. Koppejan et al.~\cite{koppejan2011neuroevolutionary} used the neuroevolutionary approach to exploiting domain expertise to learn safe models for model-based RL. Thomas et al.~\cite{thomas2021safe}, in a different approach, used near-future imagination to plan safe trajectories ahead of time. Moldovan et al.~\cite{moldovan2012safe} focused on risk aversion in MDPs using near-optimal Chernoff bounds. Lyapunov functions have also been used to guarantee safety during training~\cite{berkenkamp2017safe, chow2018lyapunov}, though constructing Lyapunov functions remains a challenge due to their typically hand-crafted nature and the absence of clear principles for agent safety and performance optimization.

In summary, RL agent evaluation relies on human-provided rewards, but the risk of misstated human objectives is often overlooked. Also, despite significant progress in safe RL methodologies, there remains a big gap in readily deployable agents in industrial contexts. Moreover, a critical balance exists between reward and cost in RL, as each action can impact both aspects, creating a multi-dimensional problem. These challenges hinder robust and dependable RL algorithms suitable for real-world implementation. Our motivation for this work is to use metagradient optimization for self-tuning of the safety threshold. This will minimize the need for safety-related hyperparameter tuning in safe RL while improving performance. Ultimately, the self-tuning safety threshold will enable us to deploy the agent directly in the real world. 
\section{Background}

In this section, we investigate the background by exploring essential preliminary concepts that serve as the foundation for this paper. We start with the discussion of the CMDP framework. Furthermore, we delve into the formulation of the safety critic and SAC. 

\subsection{Constrained Markov Decision Process (CMDP)}\label{CMDP}
CMDP comprises the tuple $<\S,\A,\P,r,c,\gamma_r,\gamma_c, \rho_0>$ where $S$ denotes the state space, $A$ represents the action space, and $r$ denotes the reward function: $r :\S \times \A \times \S \mapsto \mathbb{R}$. The transition function $\P: \S \times \A \times \S \mapsto [0,1]$ defines the likelihood $\P(s'|s,a)$ of moving from state $s$ to $s'$ by executing action $a$. The probability distribution function $\rho_0: \S \mapsto [0,1]$ denotes initial state distribution of the framework. Furthermore, $c(s)$ is the constraint indicator function which determines whether state $s$ violates the constraint functions specified by $C$: $c(s) = \mathds{1}[C(s)==1]$. Parameters $\gamma_r \in [0,1)$ and $\gamma_c \in [0,1)$ serve as discount factors for reward and safety critics, respectively. Ultimately, the solution to CMDP is represented by the policy $\pi: \S\times\A \mapsto [0,1]$ which is the probability distribution over actions. The value function associated with policy $\pi$ for a specific state-action pair $(s,a)$ and the corresponding recursive equation, known as the Bellman equation, can be formulated as follows:
\begin{equation}\label{bellman}
\begin{aligned}
    &Q^{\pi}_{r}(s,a) = \mathbb{E}_{s_t\sim \P,a_t\sim\pi}[\sum_{t=0}^\infty \gamma^{t} r(s_t,a_t) |s_0 = s, a_0=a]\\
    &\qquad\qquad=\mathbb{E}^\pi_{s'\sim \P}[r(s,a)+\gamma V^{\pi}_{r}(s')]
\end{aligned}
\end{equation}
Additionally, the primary function of the safety critic is to estimate the probability of a policy failure occurring in the future, determined by the expected cumulative discounted probability of failure.
\begin{equation}\label{safebell}
\begin{aligned}
    &Q_{c}^{\pi}(s,a) = \mathbb{E}_{s_t \sim \P, a_t \sim \pi}\big[c(s) + (1 - c(s))\sum_{t = 1}^{\infty} [\gamma_{c} ^ t c(s_t)]\big] \\
    & ~~~~~~~~~~= \text{Pr}[c(s) == 1] + \gamma_{c}\mathbb{E}_{s' \sim \P}\big[(1 - c(s))V_{c}^\pi(s')\big] 
\end{aligned}
\end{equation}
Finally, the main objective of an RL algorithm in a CMDP framework is to find a policy to maximize expected return while satisfying the constraints starting from the initial state $s_0$:
\begin{equation}
\begin{aligned}
    &\pi^*=\underset{\pi \in \Pi}{\text{argmax}} ~\mathcal{J}^\pi_r = \underset{\pi \in \Pi}{\text{argmax}} ~\mathbb{E}^{\pi}_{s_0 \sim \rho_0}[\sum_{t=0}^{\infty}\gamma ^t r_t]
    \\
    &~~~~~~~~~~~\mathrm{s.t.}~~\mathcal{J}_c^{\pi}=~\mathbb{E}^{\pi}_{s_0 \sim \rho_0}[\sum_{t=0}^{\infty}\gamma_c ^t c_t]\leq \varepsilon
\end{aligned}
\end{equation}

\subsection{Soft Actor Critic (SAC)}

SAC~\cite{haarnoja2018soft} optimizes a stochastic policy in an off-policy manner, utilizing two neural networks: one for estimating $Q$-function (critic) and another for policy updates (actor). A key feature of SAC is entropy regularization, where the policy aims to strike a balance between maximizing expected return and maximizing entropy. This balance mirrors the exploration-exploitation trade-off; higher entropy encourages greater exploration, potentially accelerating learning and prevent convergence to suboptimal solutions.

Considering $\omega_r$ and $\phi$ as parameters representing the critic and actor networks, respectively, training these networks involves sampling a batch of samples from the replay buffer. $\omega_r$ is updated by taking the gradient through the mean squared error (MSE) loss between the critic output and the target value:

\begin{equation}\label{criticsac}
\begin{aligned}
    &\mathcal{J}_r^{Q_{\omega_r}} = \mathbb{E}_{(s,a,r)\sim\mathcal{D}}
    [\frac{1}{2}(Q_{\omega_r}(s,a) - Q_{r}^{\text{tar}}(s,a))^2], 
\end{aligned}
\end{equation}
where $Q^{\mathrm{tar}}_r$ is calculated as:
\begin{multline}\label{target}
    Q_{r}^{\text{tar}}(s,a) = \\ \mathbb{E}_{\substack{s'\sim \P(s,a)\\a'\sim\pi_{\phi}}}[r(s,a) + \gamma_r(Q_r(s',a') - \alpha log(\pi_\phi(a'|s')))] 
\end{multline}
Furthermore, the policy $\pi_\phi$ is optimized by taking the gradient through the critic and the expected entropy of the policy:
\begin{equation}
    \mathcal{J}_r^{\pi_{\phi}} = \mathbb{E}_{\substack{s\sim\mathcal{D}\\a\sim\pi_{\phi}}}[\alpha log(\pi_\phi(a|s)) - Q_{\omega_r}(s,a)]
\end{equation}
Finally, it is important to note that the safety critic ($Q_{\omega_c}$) defined in Section~\ref{CMDP} is trained using the same loss formulation in Eq.~\ref{criticsac}, without the entropy term.

\begin{figure}[t]
    \centering    
    \includegraphics[trim={0.cm 6cm 0cm 0.cm},clip,width=0.7\columnwidth]{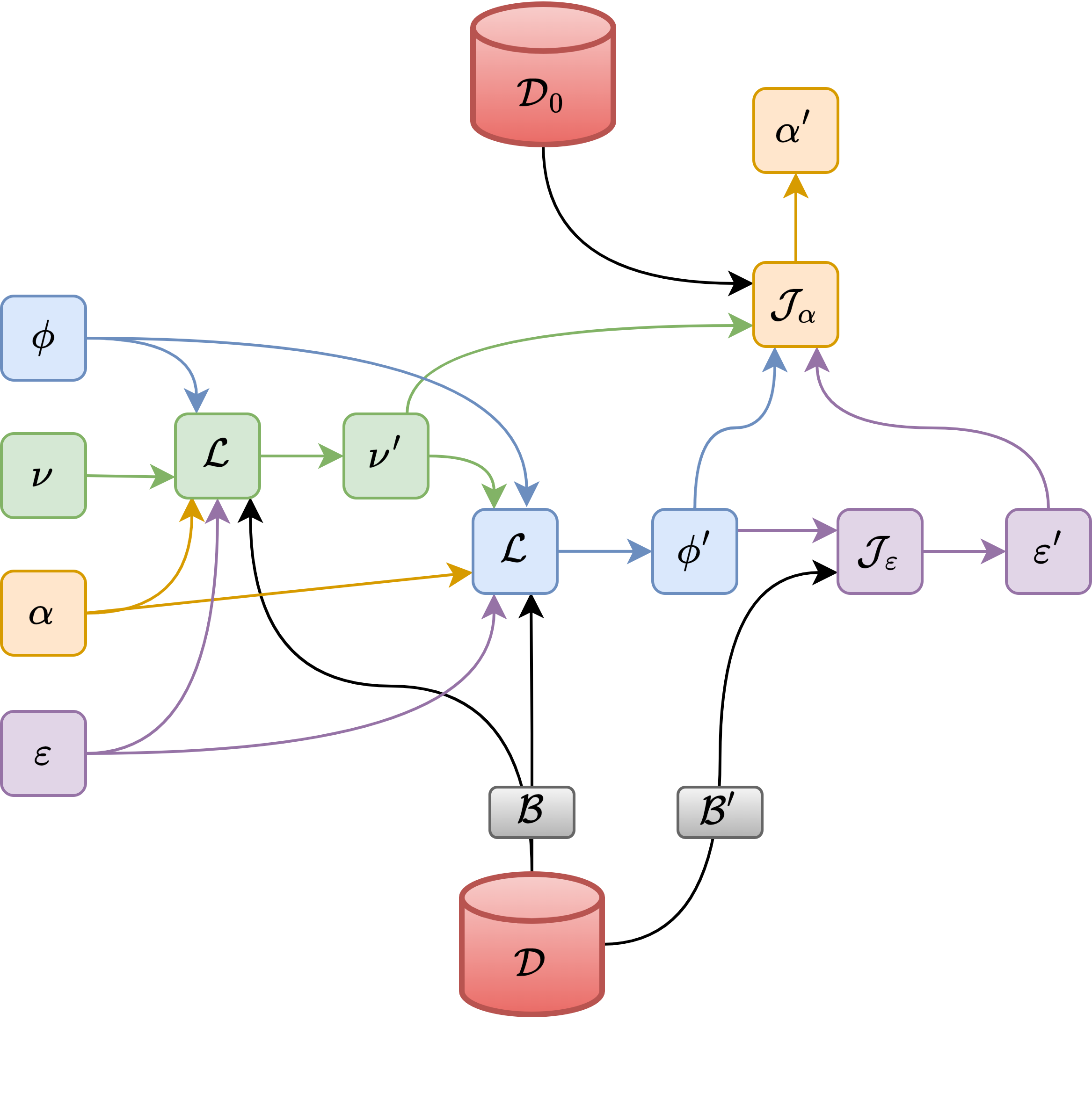}
    \caption{The Computational Graph of the Meta SAC-Lag.}
    \label{metagraph}
\vspace{-1.75em}
\end{figure}

\section{Method}
In this section the process of metagradient optimization of the safety threshold $\varepsilon$ and entropy temperature $\alpha$ is discussed.
\subsection{Metagradient Optimization}\label{metagrad}
Metagradient optimization is the process with which we can optimize the hyperparameters that are not a part of the main loss function. Fundamentally, these meta-parameters\footnote{In this paper, we use hyperparameters and meta-parameters terms interchangeably.} dictate the dynamics of the system and direct it toward a certain behavior. In the context of metagradient reinforcement learning~\cite{xu2018meta}, in abstract terms, the learnable system variables are parameterized as~$\theta$. These parameters are updated to $\theta'$ by following the rule:
\begin{equation}\label{metagradupdate}
    \theta'=\theta+f(\mathcal{J},\theta,\eta,\mathcal{B})
\end{equation}
where $\eta$ is the list of hyperparameters, $\mathcal{B}$ a mini-batch of experience, and $f$ the gradient of the objective function $\mathcal{J}$ w.r.t. $\theta$.
Furthermore, the optimization process of the meta-parameters $\eta$ can be formulated based on the updated parameter~$\theta'$:
\begin{equation}
    \eta'=\eta+\beta_\eta\frac{\partial\mathcal{J}'(\theta',\eta,\mathcal{B}')}{\partial\eta}= \eta+\beta_\eta\frac{\partial\mathcal{J}'(\theta',\eta,\mathcal{B}')}{\partial\theta'}\frac{\mathrm{d}\theta'}{\mathrm{d}\eta}
\end{equation}
where $\mathcal{J}'$ is the meta-objective used for the optimization of the meta-parameters, $\beta_\eta$ the learning rate associated with $\eta$, and $\mathcal{B}'$ a resampled mini-batch validation set similar to the cross-validation method in the meta-optimization literature~\cite{beirami2017optimal}.
Finally, $\frac{\mathrm{d}\theta'}{\mathrm{d}\eta}$ can be calculated as:
\begin{equation}
    \frac{\mathrm{d}\theta'}{\mathrm{d}\eta}=\biggl(I+\frac{\partial f(\mathcal{J},\theta,\eta,\mathcal{B})}{\partial\theta}\biggr)\frac{\mathrm{d}\theta}{\mathrm{d}\eta}+\frac{\partial f(\mathcal{J},\theta,\eta,\mathcal{B})}{\partial\eta}
\end{equation}

\subsection{SAC-Lagrangian}
In the Lagrangian version of the SAC we aim to optimize the policy based on its reward objective such that it is compliant with the safety objective:
\begin{equation}\label{optim}
\begin{aligned}
    &\pi^*_{\phi}=
    \max_{\pi_{\phi}\in\Pi}
    ~\mathcal{J}_{r_{\pi_{\phi}}}^{\pi_{\phi}}=\mathbb{E}_{\substack{s\sim\mathcal{D}\\a\sim\pi_{\phi}}}[Q_{\omega_r}(s,a)-\alpha\log{\pi_\phi(s,a)}]\\
    &~~~~~~~~~~~~~~~~\mathrm{s.t.}~~\mathcal{J}_c^{\pi_{\phi}}=\mathbb{E}_{\substack{s\sim\mathcal{D}\\a\sim\pi_{\phi}}}[Q_{\omega_c}(s,a)]\leq \varepsilon
\end{aligned}
\end{equation}
Naturally, multiple constraints can be defined for the policy to consider all of them. However, in this paper, in order to keep the formulation simple and general, we consider a single constraint signal that is the result of the superposition of all the constraint functions. In this paper, in contrast to~\cite{haarnoja2018soft}, we refrain from considering $\alpha$ as an additional constraint and aim to optimize it through metagradient optimization. 

Furthermore, the optimization process of policy in Eq.~\ref{optim} is formulated by casting it as a Lagrangian loss and backpropagating through the loss:
\begin{equation}\label{lagrange}
\begin{aligned}
    &\min_{\nu\geq0}~\max_{\pi_{\phi}\in\Pi}~\mathcal{L}(\pi_{\phi},\nu,\varepsilon,\alpha)=\mathcal{J}_{r_{\pi_{\phi}}}^{\pi_{\phi}}-\nu(\mathcal{J}_c^{\pi_{\phi}}-\varepsilon)\\
    &~~~~=\mathbb{E}_{\substack{s\sim\mathcal{D}\\a\sim\pi_{\phi}}}[Q_{\omega_r}(s,a)-\alpha\log{\pi_\phi(s,a)}-\nu(Q_{\omega_c}(s,a)-\varepsilon)]
\end{aligned}
\end{equation}
where $\nu$ is the Lagrange multiplier.

\subsection{Meta SAC-Lag}\label{metasac}
Following the conventional notation in the context of gradient-based hyperparameter optimization~\cite{franceschi2018bilevel}, we split the parameters into \textit{inner} and \textit{outer} parameters. 
Rather than a one-shot optimization as in Eq.~\ref{metagradupdate}, we propose a sequential updating approach. We define and update the inner parameters as:
\begin{equation}
    \theta'_{\mathrm{inner}}=
    \begin{bmatrix}
        \nu' \\
        \phi' 
    \end{bmatrix}=
    \begin{bmatrix}
        \nu \\
        \phi
    \end{bmatrix}+
    \begin{bmatrix}
        -\nabla_\nu \mathcal{L}(\pi_{\phi},\nu,\varepsilon,\alpha)\\
        \nabla_\phi \mathcal{L}(\pi_{\phi},\nu',\varepsilon,\alpha)
    \end{bmatrix}
\end{equation}

Furthermore, in the same sequential manner, we first update $\varepsilon$ and then $\alpha$:
\begin{equation}
    \theta'_{\mathrm{outer}}=
    \begin{bmatrix}
        \varepsilon' \\
        \alpha'
    \end{bmatrix}=
    \begin{bmatrix}
        \varepsilon \\
        \alpha
    \end{bmatrix}
    +\begin{bmatrix}
        \nabla_\varepsilon\mathcal{J}_\varepsilon(\pi_{\phi'})
        \\
        \nabla_\alpha\mathcal{J}_\alpha(\pi_{\phi'},\nu',\varepsilon')
    \end{bmatrix}
\end{equation}
where $\mathcal{J}_\varepsilon$ and $\mathcal{J}_\alpha$ correspond to the objective functions of $\varepsilon$ and $\alpha$, respectively.
To this end, we intended to design the objective function for $\varepsilon$ solely based on the performance of the resultant policy. Our intuition behind the aforementioned design stems from the idea that the threshold should be adjusted such that it improves the performance of the agent as a whole. For that purpose, the $\varepsilon$ objective function is proposed as:
\begin{equation}\label{j2eps}
    \mathcal{J}_\varepsilon(\pi_{\phi'})=
     \mathbb{E}_{\substack{s\sim\mathcal{D}\\a\sim\pi_{\phi'}}}[\nu_{\mathrm{copy}}'Q_{\omega_c}(s,a)-Q_{\omega_r}(s,a)]
\end{equation}

\begin{algorithm}[t]
\caption{Meta SAC-Lag}\label{pseudo}
\begin{algorithmic}[1]
\Require 
\Statex Initialize Policy network $\phi^0$, Exploration rate $\alpha^0$
\Statex Critic network $\omega^0_{r_1},\omega^0_{r_2}$, Safety critic network $\omega^0_{c_1},\omega^0_{c_2}$
\Statex Lagrangian values $\varepsilon^0,\nu^0$
\Statex Learning rates $\beta_\phi,\beta_\varepsilon,\beta_\nu,\beta_\alpha$
\State Create Transition buffer $\mathcal{D}$, Safety buffer $\mathcal{D}_\mathrm{s}$, and Initial state buffer $\mathcal{D}_0$
\State Randomly sample initial state $s_0\sim \rho_0$ and fill $\mathcal{D}_0$
\For{$e=1,\ldots$}
    \State Reset environment $s_0\sim\rho_0=env.reset()$
    \For{$t=0,\ldots,T-1$}
        \State Sample action $a_t\sim\pi_{\phi}$
        \State $s_{t+1},r_t,c_t\leftarrow env.step(a_t)$
        \If{$c_t==1$}
            \State $\mathcal{D}_s\leftarrow\mathcal{D}_s\cup(s_t,a_t,c_t,s_{t+1})$
        \Else
            \State $\mathcal{D}\leftarrow\mathcal{D}\cup(s_t,a_t,r_t,c_t,s_{t+1})$
        \EndIf
        \State Train $\omega_{c_1},\omega_{c_2}$ on $\mathcal{D}\cup\mathcal{D}_s$ (Eq.~\ref{safebell})
        \State Sample a batch of transitions $\mathcal{B}=$
        \Statex \qquad\qquad\qquad\qquad\qquad\qquad\qquad $\{(s,a,r,c,s')\}\in\mathcal{D}$
        \State Train $\omega_{r_1},\omega_{r_2}$ using $\mathcal{B}$ (Eq.~\ref{criticsac})
        \State $\nu'\leftarrow\nu-\beta_\nu\nabla_\nu\mathcal{L}(\pi_\phi,\nu,\varepsilon,\alpha)$ using $\mathcal{B}$ (Eq.~\ref{lagrange})
        \State $\phi'\leftarrow\phi+ \beta_\phi\nabla_\phi\mathcal{L}(\pi_\phi,\nu',\varepsilon,\alpha)$ using $\mathcal{B}$ (Eq.~\ref{lagrange})
        \State Resample $\mathcal{B'}=\{s\in\mathcal{D}\}$
        \State $\varepsilon'\leftarrow\varepsilon+\beta_\varepsilon\nabla_\varepsilon\mathcal{J}_\varepsilon(\pi_{\phi'})$ using $\mathcal{B}'$ (Eq.~\ref{j2eps})
        \State $\alpha'\leftarrow\alpha+\beta_\alpha\nabla_\alpha\mathcal{J}_\alpha(\pi_{\phi'},\nu',\varepsilon')$ using $\mathcal{D}_0$ (Eq.~\ref{jalpha})
        \State $\nu\leftarrow\nu',\phi\leftarrow\phi',\varepsilon\leftarrow\varepsilon',\alpha\leftarrow\alpha'$
        \If{$c_t==1$} Break
        \EndIf
    \EndFor
\EndFor
\end{algorithmic}
\end{algorithm}

The objective function $\mathcal{J}_\varepsilon$ is consistent with the objective function of the policy. This is evident by comparing Eq.~\ref{j2eps} with the gradient w.r.t. the policy parameters $\phi$ in Eq.~\ref{lagrange}. The objective function for $\varepsilon$ is designed to minimize the policy objective. This design stems from the idea that $\varepsilon$ aims to capture the worst-case performance of the policy $\pi_{\phi'}$. Hence, by being optimized in this way, the safety region of the policy can be correctly adjusted to reflect that. It is important to note that we use $\nu_\mathrm{copy}'$ to indicate that we merely use $\nu'$ value in the objective function and not include its gradient w.r.t. $\varepsilon$. We observed better performance by the gradient detachment in our early experiments which may be due to the injection of bias in $\nu'$ into its optimization process.
Furthermore, to optimize the exploration value $\alpha$, \cite{wang2020meta} used $Q_{\omega_r}$ as the objective function to change the value based on the performance of the policy. Therefore, in order to make the exploration rate of the Meta SAC-Lag safety compliant, we propose the objective function of $\alpha$ as:
\begin{multline}\label{jalpha}
    \mathcal{J}_\alpha(\pi_{\phi'},\nu',\varepsilon')=\\
    ~\max_{0<\alpha\leq1}\mathbb{E}_{\substack{s_0\sim\rho_0\\a\sim\pi_{\phi'}^\mathrm{det}}}[Q_{\omega_r}(s_0,a)-\nu'(Q_{\omega_c}(s_0,a)-\varepsilon')]
\end{multline}
where $\pi_{\phi'}^\mathrm{det}$ indicates the deterministic action value output by the policy. Basically, we use the expectation of the Lagrangian formulation evaluated in the initial states encountered by the agent. To gain a better understanding of the gradient relations, illustration of the optimization process of Meta SAC-Lag is depicted in Fig.~\ref{metagraph}.

\subsection{Implementation Details}
The learning process of Meta SAC-Lag is presented in Algorithm~\ref{pseudo}. 
The proposed algorithm utilizes three replay buffers for training. The main replay buffer $\mathcal{D}$ stores all the transitions occurred while interacting with the environment, safety replay buffer $\mathcal{D}_\mathrm{s}$ stores all the transitions that have led to a constraint violation, and $\mathcal{D}_0$ builds an approximation of $\rho_0$ by generating samples from the distribution. We used a sampled batch $\mathcal{B}\subset\mathcal{D}$ to train the critic networks and the \textit{inner} parameters $\nu$ and $\phi$. Following that, as discussed in Section~\ref{metagrad}, we use resampled $\mathcal{B}'\subset\mathcal{D}$ and $\mathcal{D}_0$ to train the meta-parameters $\varepsilon$ and $\alpha$, respectively. The resampling process is analogous to the meta-testing process and is used to reduce bias in the training of the outer parameters~\cite{beirami2017optimal,franceschi2018bilevel}.
Moreover, following the original architecture~\cite{haarnoja2018soft}, Meta SAC-Lag uses two critic and safety critic networks to prevent the overestimation of the value functions. To this end, the target values in Eq.~\ref{target} are calculated as $\min\{Q_{\bar{\omega}_{r_1}},Q_{\bar{\omega}_{r_2}}\}$ and $\max\{Q_{\bar{\omega}_{c_1}},Q_{\bar{\omega}_{c_2}}\}$, respectively. The $\bar\omega$ notation is used to indicate the target networks which are copies of the main networks updated with a time delay. Proposed in~\cite{lillicrap2015continuous}, the target networks aim to increase the stability of the training process and are calculated using the polyak averaging: $\bar\omega=\tau\omega+(1-\tau)\bar\omega$. The hyperparameter $\tau\in(0,1)$ typically has a value near zero.

Finally, it is also worth mentioning, in contrast to the original SAC, we use RMSProp~\cite{ruder2016overview} instead of Adam to calculate the higher-order gradients of the parameters $\nu$, $\phi$, and $\varepsilon$ since backpropagating through RMSProp seems to be more numerically stable~\cite{wang2020meta}.

\section{Experiments}
\begin{figure*}[t]
\centering
\captionsetup{justification=centering}
\includegraphics[trim={0.2cm 0.75cm 1cm 0cm},clip,width=2.1\columnwidth,height=0.40\linewidth]{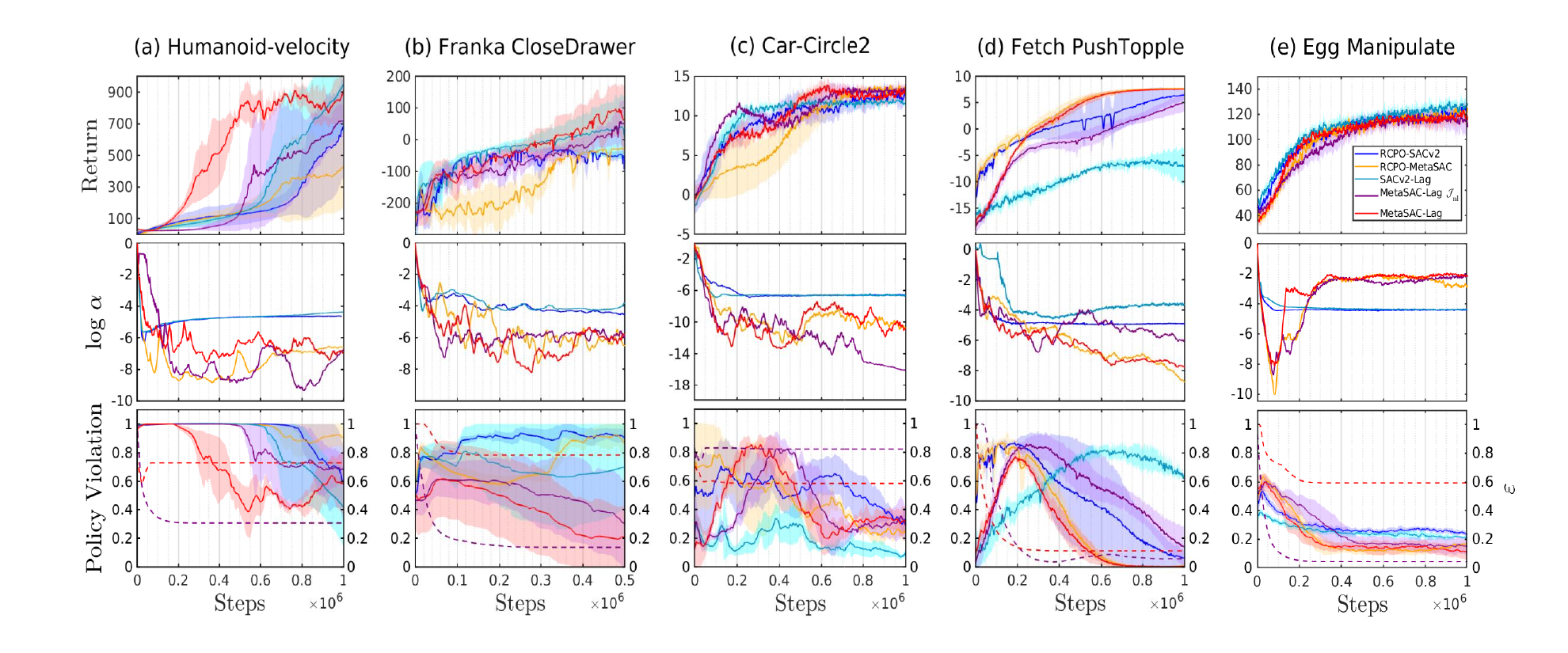}

\caption{
Performance of Meta SAC-Lag compared with the baseline algorithms. \textbf{(Top row):} Reward performance during the learning process. (Higher values are better) \textbf{(Middle row):} The value of Exploration hyperparameter ($\alpha$). \textbf{(Bottom row):} Episodic policy safety performance of the algorithms during the learning process. (Lower values are better). The dashed lines illustrate the constraint threshold value ($\varepsilon$).}
\label{sim_res}
\vspace{-1em}
\end{figure*}

In this section, we evaluate the performance of Meta SAC-Lag. Specifically, our aim is to study two questions:
\begin{itemize}
    \item How much does the added autonomy affect the performance of the algorithm compared to the baseline methods?
    \item How capable is Meta SAC-Lag to learn optimal performance in a real-world setup while avoiding actions that might catastrophically damage the system?
\end{itemize}
\subsection{Test Benchmarks and Baselines}
In order to study how the proposed algorithm will perform in safety-critical robotic scenarios, we use five simulated robotic environments with four different themes:
\begin{itemize}
    \item \textbf{Locomotion}: In this theme, the purpose of control is to move the robotic system in the forward direction. The safety constraints are violated whenever the controller's actions make the system exceed its limits, e.g., the velocity is higher than a certain threshold or the robot is falling to the ground. For that purpose, we use the Mujoco-based~\cite{todorov2012mujoco} Humanoid-Velocity environment from the Safety Gymnasium codebase~\cite{ji2024safety}. It is important to note that the safety-related reward shaping of this environment is removed to have a better understanding of the safety performance of the algorithms.
    \item \textbf{Obstacle Avoidance}: In many real-world robotic applications, there are mobile robots with manipulation capabilities. An important constraint of these systems is achieving their goal while avoiding certain regions in their surroundings. We adopt Isaac Gym-based FreightFrankaCloseDrawer~\cite{liang2018gpu}. In this setup, the robot attempts to get near a drawer and close it while avoiding a red region. In addition, we use Car-Circle2 task~\cite{ji2024safety} where the objective is to steer a car in a circular motion while avoiding collision with two walls.
    \item \textbf{Manipulation}: Another important area of safety-concerned robotic applications is manipulation. For that purpose, we use two embodied scenarios. For the \textbf{robotic manipulation} task we use Push Topple~\cite{bharadhwaj2020conservative,hsu2022improving} environment where the robotic arm must relocate a box without toppling it. Furthermore, in the \textbf{dexterous manipulation} scenario, we adopt the Egg Manipulate task where the agent must rotate an egg to a specific orientation without dropping it or exerting a force of more than 20 N. For both tasks, we use the Gymnasium Robotics codebase~\cite{gymnasium_robotics2023github}.
\end{itemize}
It is important to note that in the training process, we treat the constraints as hard constraints and terminate the episode whenever a violation has happened in the system.
Furthermore, three baseline algorithms are chosen to compare and study the performance of Meta SAC-Lag:
\begin{itemize}
    \item[-] \textbf{SACv2-Lag}: The basic form of Meta SAC-Lag which uses Eq.~\ref{lagrange} to optimize the policy and the Lagrangian multiplier with a fixed safety threshold.
    \item[-] \textbf{Reward Constrained Policy Optimization (RCPO-SACv2)}: Optimizes the policy using the $Q$-function formulated as $\mathbb{E}_\pi[\hat Q(s,a)=Q_r(s,a)-\nu Q_c(s,a)]$. The dual variable $\nu$ is also updated using Eq.~\ref{lagrange}.
    \item[-] \textbf{RCPO-MetaSAC}: To show the effectiveness of our safe exploration technique, we use the $\hat Q(s,a)$ formulation in RCPO and optimize $\alpha$ using the approach proposed in~\cite{wang2020meta}.
    \item [-] \textbf{Meta SAC-Lag $\mathcal{J}_{nl}$}: Inspired by~\cite{honari2024safety}, we experiment with a nonlinear objective function for $\varepsilon$ specified as:
    \begin{multline}\label{j1eps}
    \mathcal{J}_\varepsilon^{nl}(\pi_{\phi'})= \\ ~~~~\mathbb{E}_{\substack{s\sim\mathcal{D}\\a\sim\pi_{\phi'}}}
    \biggl[\scalemath{0.8}{
    \begin{cases}
        Q_{\omega_r}(s,a)Q_{\omega_c}(s,a) & \mathrm{if}~Q_{\omega_r}(s,a)<0 \\
        Q_{\omega_r}(s,a)(1-Q_{\omega_c}(s,a)) &  \mathrm{otherwise}\\
    \end{cases}
    }\biggr]
    \end{multline}
Essentially, $\mathcal{J}_\varepsilon^{nl}$ can have the advantage of no reliance on external parameter values, as opposed to Eq.~\ref{j2eps} which uses $\nu'$ in the objective formulation.
\end{itemize}
To have a fair comparison, we tune the values of $\varepsilon$ and $\nu$ for SACv2-Lag and RCPO-SACv2. Also, we use the values of RCPO-SACv2 for RCPO-MetaSAC. The values are outlined in Table~\ref{hyperparams}. For the value of $\alpha$, SACv2 constrains the policy entropy as $\mathbb{E}_{\substack{s\sim\mathcal{D}\\a\sim\pi}}[-\log(\pi(s_t,a_t))]\geq \mathcal{H}$ and defines the $\alpha$ loss as $\min_{\alpha>0} \mathcal{L}(\alpha)=\mathbb{E}_{\substack{s\sim\mathcal{D}\\a\sim\pi}}[\alpha(\log(\pi(s_t,a_t))+\mathcal{H})]$. The authors propose the formula $\mathcal{H}=-\mathrm{dim}(\mathcal{A})$ as their target entropy. Furthermore, two important initial hyperparameter values of Meta SAC-Lag are automatically tuned; therefore, we set $\varepsilon=1$ and $\alpha=1$ as their initial values. Moreover, due to their similar training pipelines, we use the initial values of $\nu$ for SACv2-Lag in Table~\ref{hyperparams} for Meta SAC-Lag. We also set $\gamma_r=0.99$ and $\gamma_c=0.6$ for all the tasks. The results indicate the mean and variance of the performance of the algorithms across multiple independent runs.

\begin{table}[t]
  \centering
  \captionsetup{justification=centering}
\caption{Hyperparameter values ($\varepsilon$ and $\nu$) of the comparison methods}
\label{hyperparams}
\begin{adjustbox}{max width=\columnwidth}
\begin{tabular}{cccc}
\hline & \\[-1.75ex]
\textbf{Environment / Parameter} & $\varepsilon$ & \begin{tabular}[c]{@{}c@{}}Meta SAC-Lag \\  SACv2-Lag\end{tabular} & \begin{tabular}[c]{@{}c@{}}RCPO-SACv2 \\ RCPO-MetaSAC\end{tabular} \\ \hline & \\[-1.75ex]
Humanoid-Velocity& 0.4& 10& 10\\
Franka DrawerClose& 0.6& 10&10\\
Car-Circle2& 0.5& 100& 1\\
Fetch PushTopple&0.5&1000&10\\
Egg Manipulate&0.5&100&1\\ \hline & \\[-2ex]
\end{tabular}
\end{adjustbox}
\vspace{-1.5em}
\end{table}

\subsection{Simulation Results}
The simulation results are depicted in Fig.~\ref{sim_res}. To this end, we also report the return and the policy episodic violation rate in Table~\ref{results_table}. The violation rate is calculated as the average number of failures over a specific window of episodes. The results not only indicate that Meta SAC-Lag provides automated tuning of the safety-related hyperparameters but also, that the convergence process of the policy incurs lower constraint violations and yields higher or comparable returns. Furthermore, the update profile of $\alpha$ shows that as training goes on, in most cases, Meta SAC-Lag updates $\alpha$ to values lower than SACv2. This indicates that as the policy converges to a near-safe optimal solution, $\alpha$ is rapidly decreased to favor exploitation and prevent further constraint violations. Moreover, we can observe similar $\alpha$ profiles in Meta SAC-Lag and RCPO-SACv2 which can be attributed to $\alpha$ being optimized using similar objective functions. In addition, the optimization process of $\varepsilon$ shows a generally fast convergence. The fast convergence of $\varepsilon$ provides the advantage of stable optimization as other values can updated based on the optimally achieved value of $\varepsilon$. Finally, regarding the comparison between Eq.~\ref{j2eps} and Eq.~\ref{j1eps} we observe consistently better performance of Eq.~\ref{j2eps} in both aspects of return and safety. In summary, the optimization outcomes of Meta SAC-Lag demonstrate that the algorithm excels across a range of embodied control tasks, proficiently learning optimal solutions, while demanding minimal hyperparameter tuning.

\begin{table*}[h]
\caption{Relative Performance of the Algorithms During the Learning Process \\(The best performance is shown in bold)}
\label{results_table}
\centering
\begin{tabular}{ccccccccccc}
\hline & \\[-1.75ex]
\multirow{2}{*}{\textbf{Task / Method}} & \multicolumn{2}{c}{SACv2-Lag} & \multicolumn{2}{c}{RCPO-SACv2} & \multicolumn{2}{c}{RCPO-MetaSAC} & \multicolumn{2}{c}{MetaSAC-Lag} & \multicolumn{2}{c}{MetaSAC-Lag $\mathcal{J}_\mathrm{nl}$} \\& Jr& Jc& Jr& Jc& Jr& Jc& Jr& Jc& Jr& Jc\\ \hline & \\[-1.75ex]
Humanoid-Velocity& \textbf{956.11}& \textbf{0.42}& 690.66& 0.57& 435.93& 0.91& 812.90& 0.49& 712.72& 0.67\\
Franka DrawerClose& 2.19& 0.69& -67.20& 0.89& -29.85& 0.85& \textbf{63.13}& \textbf{0.20}& 54.02& 0.30\\
Car-Circle2&11.65&\textbf{0.10}&12.67&0.32&12.36&0.24&13.79&0.31&\textbf{13.95}&0.29\\
Fetch-Topple& -6.6& 0.66& 6.55& 0.06& \textbf{7.6}& 0.007& 7.49& \textbf{0.004}& 4.98& 0.17\\
Egg Manipulate& \textbf{128.66}& 0.21& 121.53& 0.23& 124.99& 0.17& 115.44& \textbf{0.11}& 109.95& 0.14\\ \hline
\end{tabular}
\vspace{-0.5em}
\end{table*}

\subsection{Real-World Deployment}
Deployability can be regarded as one of the most important obstacles in using RL for learning to control real-world systems~\cite{enayati2023}. Choosing unsafe actions might lead the system to states that might damage it catastrophically, if chosen repeatedly. Therefore, using the conventional safe RL algorithms hinders their deployability since they require intensive hyperparameter tuning. In line with our purpose of assessing the deployability of a safe RL method, we propose a simple, yet important, safe RL testbench. This task, which we call \textit{Pour Coffee}, is the task of moving a coffee-filled mug from a home position to a specific location and pouring the coffee into another cup. The task is executed using a Kinova~Gen3 robot and its digital twin is created in the PyBullet simulation environment~\cite{coumans2021}. We define the state space 
$\mathcal{S}=\biggl\{X_{cup}\,\cup~O_{cup}~\cup~\dot X_{cup}~\cup~X_{goal}~\cup~O_{goal}\biggr\}$
where $X=\{x,y,z\}$ and $O=\{\psi,\theta,\phi\}$ refer to the Cartesian position and the Euler angles in the Tait-Bryan ZYX intrinsic convention, respectively. Furthermore, the action of the agent maps to the velocity of the end-effector: $\mathcal{A}=\{\dot x_{cup},\dot y_{cup},\dot z_{cup},\dot \phi_{cup}\}$. Moreover, we hierarchically define the reward function for reaching and pouring the coffee based on the Euclidean distance between the cup and the goal $d=||X_{cup}-X_{goal}||_2$:
\begin{equation}
\scalemath{0.88}{
    r(s,a,s')=
    \begin{cases}
        r_1\cdot d+r_2\cdot||\ddot X_{cup}||+r_3\cdot\mathds{1}[\mathrm{spillage}] &\mathrm{if}~ d>d_{\mathrm{thresh}}\\
        -|\phi_{cup}-\phi_{goal}|+10 &\mathrm{otherwise}
    \end{cases}
}
\end{equation}

where $r_1=-2,~r_2=-0.05,~r_3=-1$ and $d_{\mathrm{thresh}}=5~cm$. Furthermore, the system violates the safety constraints whenever self-collision or collision with the environment objects occurs. Additionally, we can define another constraint as spilling the coffee. As will be shown, this constraint forces the policy to be less jerky and aims to minimize the acceleration. The advantage of this approach, in contrast to similar environments~\cite{zhu2020robosuite}, is the fact that it will eliminate the need to engineer the reward function to minimize the jerk and acceleration of the robot.

\begin{table}[t]
\caption{\textit{Pour Coffee} Reward-Constraint Settings}\label{pour_setting}
\centering
\begin{adjustbox}{max width=\columnwidth}
\begin{tabular}{lccccc}
\hline \\[-1.75ex]
\multicolumn{1}{c}{\multirow{2}{*}{Experiment Setting}} & \multicolumn{3}{c}{Reward} & \multicolumn{2}{c}{Violation} \\[-0.25ex] \cmidrule(lr){2-4} \cmidrule(lr){5-6} \\[-2.5ex]
\multicolumn{1}{c}{} & Distance ($r_1$) & Acceleration ($r_2$) & Penalty ($r_3$) & Collision      & Spillage \\[0.5ex]
\hline \\[-1.5ex]
Simulation \#1 & \cmark & \xmark & \xmark & \cmark & \cmark \\
Simulation \#2 & \cmark & \cmark & \xmark & \cmark & \xmark \\
Simulation \#3 & \cmark & \cmark & \xmark & \cmark & \cmark \\
Simulation \#4 & \cmark & \cmark & \cmark & \cmark & \cmark \\
Real & \cmark & \xmark & \xmark & \cmark & \cmark \\
\hline
\end{tabular}
\end{adjustbox}
\vspace{-1.5em}
\end{table}

We conduct experiments with Meta SAC-Lag in four reward and constraint settings. The experiments aim to study whether formulating the problem sub-objectives can be more practical by defining them as constraints rather than shaping the reward explicitly. In the presented task, coffee spillage provides an implicit sub-objective that can be explicitly modeled as the sub-objective of minimizing the jerk and acceleration of the end-effector during the execution of the task. As shown in Table~\ref{pour_setting}, three experiments (Simulation \#2, \#3, \#4) utilize different reward shaping schemes along with various constraint definitions. Moreover, we trained Meta SAC-Lag without engineered reward shaping (Simulation \#1) both in the simulation environment and the real-world Kinova Gen3 setup. 
In order to make comparisons and evaluate the Sim2Real capability, the simulation-trained models were deployed on the robot using the checkpoints saved during the learning process. 



\begin{figure}[t]
    \centering
    \captionsetup{justification=centering}
    \includegraphics[trim={5.5cm 1.2cm 5.5cm 0.5cm},clip,width=0.91\columnwidth]{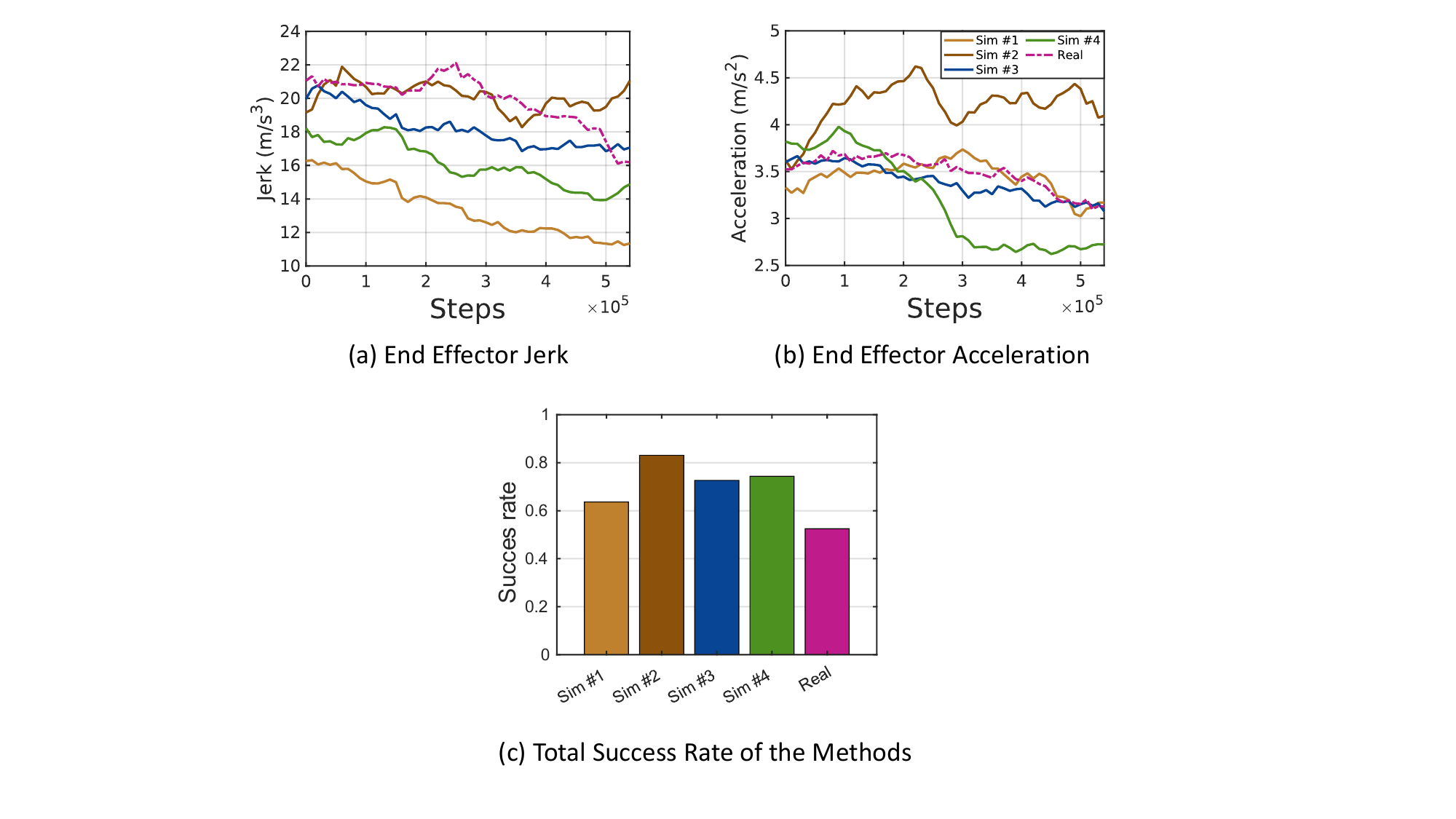}
    \caption{Deployment results of Meta SAC-Lag on the real-world setup. (a) and (b) represent the jerk and acceleration of the end effector during the training process. (c) shows the final success rate of the algorithms.}
    \label{real_res}
\vspace{-2.em}
\end{figure}

The evaluation results are depicted in Fig.~\ref{real_res}. The results illustrate that, as a result of providing a denser reward signal, explicit reward shaping can have positive effects in the increase of the success rate. However, using the spillage constraint helps the algorithm be even more effort-compliant resulting in lower jerk and comparable acceleration results. In other words, while being successful in executing the task is the most important metric, in a real-world scenario, sacrificing the performance to lower the effort of the system and satisfy other safety concerns can be reasonable. In addition, regarding the comparison between Sim2Real and Real deployment of the proposed algorithm, we can observe that while both setups have similar behaviors, the real-world deployment is slightly hindered by the system's physical limitations, such as sensor noise, control saturation, system fatigue, etc. Despite all that, the algorithm trained on the real-world setup without engineered reward function achieves results comparable to the models trained in the simulation.

\section{Conclusions}
The paper focused on the problem of automatic hyperparameter tuning in Lagrangian safe RL methods. A novel model-free architecture called Meta SAC-Lag was proposed which addressed two inherent problems: safe exploration and constraint bound tuning. To this end, through the use of metagradient optimization, the algorithm is capable of adjusting the safety-related hyperparameters with minimal initial tuning. Furthermore, we studied the performance of our algorithm in five simulated embodied applications with the themes of locomotion, obstacle avoidance, robotic manipulation, and dexterous manipulation. We observed that the synergy created between the parameters and the hyperparameters results in comparable or better performance of the policy in terms of reward or safety.
Additionally, we conducted an experiment in a real-world setup involving a practical coffee-pouring robotic environment without any explicit safety-related reward shaping. We deployed the algorithm on the Kinova Gen3 robot and showed that the proposed algorithm can be helpful for real-world safety-sensitive applications by reducing the reliance on heuristic implementation of safety. We also observed that formulation of safety solely as the violation rather than engineering the reward function results in applying lower levels of effort at the cost of a diminished success performance. This trade-off can be especially favorable in real-world setups where safety violations are costly. Specifically, the proposed algorithm will learn the optimal policy in the real-world setup while adhering to the collision constraints and minimizing the effort imposed on the robot.

\bibliographystyle{IEEEtran}

\onecolumn
\appendix

\subsection{Theoretical Analysis}
In this section, the gradients of the each component of the algorithm is derived. While the automated differentiation tools for deep learning such as PyTorch and TensorFlow provide automated gradient calculation of this algorithm, the gradient analysis of Meta SAC-Lag may provide insightful information. 

\textbf{Step 1}: In the beginning of optimization, the gradient of the Lagrangian multiplier $\nu$ is calculated using the inner loss (for ease of presentation, the expected values and the source of $s$ are dropped in this analysis):
\begin{equation}
\begin{aligned}
    \nabla_\nu\J_\nu&=\nabla_\nu\L(\pi_\phi,\nu,\varepsilon,\alpha)\\
    &=\nabla_\nu\left[Q_{\omega_r}(s,\pi_\phi(s))-\nu(Q_{\omega_c}(s,\piPHI)-\varepsilon)-\alpha\logpi\right]\\
    &=-[Q_{\omega_c}(s,\piPHI)-\varepsilon]
\end{aligned}
\end{equation}

Hence, $\nu$ is updated as:

\begin{equation}
    \nu'\gets\nu-\beta_\nu\nabla_\nu\J_\nu=\nu+\beta_\nu\left[Q_{\omega_c}(s,\piPHI)-\varepsilon\right]
\end{equation}

\textbf{Step 2}: Following the Lagrangian multiplier, the gradient of the actor parameters w.r.t. the Lagrangian loss is calculated as:

\begin{equation}
\begin{aligned}
    \nabla_\phi\J_\phi&=\L(\pi_\phi,\nu',\varepsilon,\alpha)\\
    &=\nabla_\phi\left[\Qr(s,\piPHI)-\nu'(\Qc(s,\piPHI)-\varepsilon)-\alpha\logpi\right]\\
    &=\nabla_\phi\left[\Qr(s,\piPHI)-(\nu+\beta_\nu(\Qc(s,\piPHI)-\varepsilon)(\Qc(s,\piPHI)-\varepsilon)-\alpha\logpi\right]\\
    &=\nabla_\phi \Qr(s,\piPHI)-\nabla_\phi \Qc(s,\piPHI)\left[2\beta_\nu\Qc(s,\piPHI)-2\beta_\nu\varepsilon+\nu\right]-\alpha\nabla_\phi\logpi\\
\end{aligned}
\end{equation}

The actor parameters are then updated as:
\begin{equation}
\begin{aligned}
    \phi'&\gets\phi+\beta_\phi\nabla_\phi\J_\phi\\
    &=\phi+\beta_\phi\big[\nabla_\phi \Qr(s,\piPHI)-\nabla_\phi \Qc(s,\piPHI)\left[2\beta_\nu\Qc(s,\piPHI)-2\beta_\nu\varepsilon+\nu\right]\\
    &\qquad\qquad\qquad\qquad\qquad\quad
    -\alpha\nabla_\phi\logpi\big]
\end{aligned}
\end{equation}

\textbf{Step 3}: The first meta-parameter used in the training pipeline is the safety threshold $\varepsilon$. Assuming the meta-objective formulation of Equation~\ref{j2eps}, the gradient of $\varepsilon$ can be derived as:

\begin{equation}\label{epsgrad}
\begin{aligned}
     \nabla_\varepsilon\J_\varepsilon&= \nabla_\varepsilon\left[\nu'_{\text{copy}}\Qc(s,\piPHII)-\Qr(s,\piPHII)\right]\\
     &=\nabla_\varepsilon{\phi'}^\textbf{T}\nabla_{\phi'}\J_\varepsilon+\nabla_\varepsilon\nu'\left[\nabla_{\nu'}{\phi'}^\textbf{T}\nabla_{\phi'}\J_\varepsilon+\colorcancelto{red}{0}{\nabla_{\nu'}\J_\varepsilon} \right]+\colorcancelto{red}{0}{\nabla_\varepsilon\J_\varepsilon}
\end{aligned}
\end{equation}

It should be noted that the chain of gradients that are calculated must be causal. For that reason, gradients such as $\nabla_{\phi'}\nu'$ would be meaningless and have not been written. Moreover, each component of Equation~\ref{epsgrad} can be computed as:

\begin{align}
    &\nabla_\varepsilon\phi'=2\beta_\phi\beta_\nu\nabla_\phi\Qc(s,\piPHI) \\
    &\nabla_{\phi'}\J_\varepsilon=\nu'\nabla_{\phi'}\Qc(s,\piPHII)-\nabla_{\phi'}\Qr(s,\piPHII) \label{Eq:gradJphieps}\\
    &\nabla_\varepsilon\nu'=-\beta_\nu\\
    &\nabla_{\nu'}\phi'=-\beta_\phi\nabla_\phi\Qc(s,\piPHI)
\end{align}

Therefore, the final gradient is calculated as:

\begin{equation}\label{Eq:gradeps}
\begin{aligned}
    \nabla_\varepsilon&\J_\varepsilon={\left[-3\beta_\nu\beta_\phi\nabla_\phi\Qc(s,\piPHI)\right]^\textbf{T}\left[\nabla_{\phi'}\Qr(s,\piPHII)-\nu'\nabla_{\phi'}\Qc(s,\piPHII)\right]}
\end{aligned}
\end{equation}

Hence, the value of $\varepsilon$ is updated as:

\begin{equation}
\begin{aligned}
    \varepsilon'&\gets\varepsilon+\beta_\varepsilon\nabla_\varepsilon\J_\varepsilon\\
    &=\varepsilon+\beta_\varepsilon\scalemath{1}{\left[-3\beta_\nu\beta_\phi\nabla_\phi\Qc(s,\piPHI)\right]^\textbf{T}\left[\nabla_{\phi'}\Qr(s,\piPHII)-\nu'\nabla_{\phi'}\Qc(s,\piPHII)\right]}
\end{aligned}
\end{equation}

The meta-gradient of $\varepsilon$ consists of two components. While the right hand side evaluates the performance of the new actor parameter $\phi'$ by calculating the tradeoff between safety and optimality, the left hand side is the main direction driver of the gradient. 
The left hand side of the meta-gradient equation determines the policy's degree of unsafety with respect to the safety critic. The gradient of the policy w.r.t. the safety critic would determine the effect of the changes in the policy parameters on the measure of safety. 
The gradient will always try to force the policy to become safer by moving in the negative direction of the gradient.
The idea of minimizing the policy objective discussed in Section~\ref{metasac} is also clear from the observation of the gradients of $\varepsilon$ and specifically in Equation~\ref{Eq:gradJphieps} and the left hand side of Equation~\ref{Eq:gradeps}. Without the negative sign the algorithm would try to match $\varepsilon$ with the unsafety performance of the policy rather than forcing it to become safer.

\textbf{Step 4}: The final step of the optimization involves calculating the meta-gradient of the temperature $\alpha$ which is calculated as:

\begin{equation}
\begin{aligned}
    \nabla_\alpha\J_\alpha&= \nabla_\alpha[\Qr(s,\pi^{det}_{\phi'}(s))-\nu'(\Qc(s,\pi^{det}_{\phi'}(s))-\varepsilon')]\\
    &=\colorcancelto{red}{0}{\nabla_\alpha\nu'}[\nabla_{\nu'}\J_\alpha+\nabla_{\nu'}{\phi'}^\textbf{T}\nabla_{\phi'}\J_\alpha+\nabla_{\nu'}\varepsilon'\nabla_{\varepsilon'}\J_\alpha]\\
    &~~~~~~~~~~ +\nabla_\alpha{\phi'}^\textbf{T}[\nabla_{\phi'}\J_\alpha+\nabla_{\phi'}\varepsilon'\nabla_{\varepsilon'}\J_\alpha]+\colorcancelto{red}{0}{\nabla_\alpha\varepsilon'}\nabla_{\varepsilon'}\J_\alpha
\end{aligned}
\end{equation}

The components of the gradient can then be calculated as:

\begin{align}
    &\nabla_{\varepsilon'}\J_\alpha=\nu'\\
    &\nabla_\alpha\phi'=-\beta_\phi\nabla_\phi\logpi\\
    &\nabla_{\phi'} \J_\alpha=\nabla_{\phi'}\Qr(s,\pi^{det}_{\phi'}(s))-\nu'\nabla_{\phi'}\Qc(s,\pi^{det}_{\phi'}(s)) \\
    &\nabla_{\phi'}\varepsilon'=\beta_\varepsilon\scalemath{1}{\left[-3\beta_\nu\beta_\phi\nabla_\phi\Qc(s,\piPHI)\right]^\textbf{T}\left[H_{\phi'}\Qr(s,\piPHII)-\nu'H_{\phi'}\Qc(s,\piPHII)\right]}\label{Eq:gradphipepsp}
\end{align}

Therefore, the final gradient of $\alpha$ is obtained as:

\begin{equation}
\begin{aligned}
    \nabla_\alpha\J_\alpha&=-\beta_\phi\nabla_\phi\logpi^\textbf{T}[\nabla_{\phi'}\Qr(s,\pi^{det}_{\phi'}(s))-\nu'\nabla_{\phi'}\Qc(s,\pi^{det}_{\phi'}(s))\\
    &\qquad+\nu'\beta_\varepsilon\scalemath{1}{\left[-3\beta_\nu\beta_\phi\nabla_\phi\Qc(s,\piPHI)\right]^\textbf{T}\left[H_{\phi'}\Qr(s,\piPHII)-\nu'H_{\phi'}\Qc(s,\piPHII)\right]}]\\
    & =-\beta_\phi\nabla_\phi\logpi^\textbf{T}[\nabla_{\phi'}\Qr(s,\pi^{det}_{\phi'}(s))-\nu'\nabla_{\phi'}\Qc(s,\pi^{det}_{\phi'}(s))]\\
\end{aligned}
\end{equation}

The second term of the gradient ($\nabla_{\phi'}\varepsilon'\nabla_{\varepsilon'}\J_\alpha$) was dropped because in the architecture of the neural networks used to implement Meta SAC-Lag, the ReLU activation function was used. An important feature of ReLU is the fact that it has zero second-order derivative almost everywhere. Hence, the Hessian matrix in Equation~\ref{Eq:gradphipepsp} would be equal to zero.

Finally, the update rule for $\varepsilon$ follows:

\begin{equation}
\begin{aligned}
    \alpha'&\gets\alpha+\beta_\alpha\nabla_\alpha\J_\alpha\\
    &=\alpha-\beta_\alpha\beta_\phi\nabla_\phi\logpi^\textbf{T}[\nabla_{\phi'}\Qr(s,\pi^{det}_{\phi'}(s))-\nu'\nabla_{\phi'}\Qc(s,\pi^{det}_{\phi'}(s))]
\end{aligned}
\end{equation}

By observing the right hand side of $\nabla_\alpha\J_\alpha$ it can be noticed that term is equivalent to $\nabla_{\phi'}[\Qr(s,\pi^{det}_{\phi'}(s))-\nu'\Qc(s,\pi^{det}_{\phi'}(s))]$ which is similar to the critic formulation of RCPO as $\hat{Q}^\pi(s,a)=Q^\pi(s,a)-\nu Q^\pi_c(s,a)$. Hence, in essence the meta-gradient calculation of $\alpha$ in Meta SAC-Lag and RCPO-MetaSAC are the same. This observation can be confirmed by noticing the similarity in the updating profile of $\alpha$ in the simulation results in Figure~\ref{sim_res}. The profile is also similar to that of Meta SAC-Lag$\J_{nl}$ because due to the use of ReLU and the Hessian matrix becoming zero, the effect of $\J_\varepsilon$ on $\alpha$ has been eliminated and the updating procedure for both of them would be the same.

\end{document}